\documentclass[5p,times]{elsarticle}

\usepackage{amsmath,amsfonts}
\usepackage{array}
\usepackage{textcomp}
\usepackage{url}
\usepackage{verbatim}
\usepackage{graphicx}
\usepackage{microtype}
\usepackage{xcolor}
\usepackage{booktabs}
\usepackage{multirow}
\usepackage{diagbox}
\usepackage{scalerel}
\usepackage{tikz}
\usepackage{siunitx}
\usetikzlibrary{svg.path}
\usepackage[percent]{overpic}

\usepackage{stfloats}            
\journal{Optics \& Laser Technology}

\begin{document}

\begin{frontmatter}

\title{Spatially Coupled Phase-to-Depth Calibration for Fringe Projection Profilometry}

\author{Sehoon Tak}
\author{Jae-Sang Hyun\corref{cor1}}
\ead{hyun.jaesang@yonsei.ac.kr}
\cortext[cor1]{Corresponding author}
\address{Department of Mechanical Engineering, Yonsei University, Seoul 03722, South Korea.}

\begin{abstract}
In fringe projection profilometry (FPP), depth is commonly recovered by fitting a phase-to-depth relation independently at each camera pixel. Although such pixel-wise calibration achieves high local accuracy, neighboring pixels can acquire markedly different calibration functions even when they observe the same smooth surface, producing spatially inconsistent geometry and structured surface artifacts. We propose a spatially coupled phase--depth transformation in which all pixels share a single low-dimensional mapping---global phase scalars combined with affine spatial terms on the undistorted reference-camera grid---rather than independent per-pixel fits, optionally augmented by a bounded, spatially smooth correction field. We further introduce a native-grid pairing scheme that constructs phase--depth calibration pairs directly on the reference-camera grid: when depth supervision comes from a rectified active-stereo pipeline, planes are fitted in stereo 3D and sampled back onto the camera grid along native rays, so the phase maps are never rectified. On a dental target with high-resolution scanner ground truth, the proposed model attains point-to-surface RMSE comparable to an active-stereo reference (about \(12\,\mu\mathrm{m}\) aggregate) while substantially improving spatial coherence over pixel-wise polynomial and rational calibration, and reduces the runtime mapping to a few element-wise operations per pixel with negligible parameter storage.
\end{abstract}


\end{frontmatter}

\section{Introduction}\label{sec:introduction}
Fringe projection profilometry (FPP) enables efficient, high-resolution, noncontact 3D measurement and is used in medical, industrial, and consumer applications\cite{XU2020106193, QIAN2021106382}. In a typical FPP system, a digital projector illuminates the object with structured light, and a camera records the deformed patterns. Object shape is first encoded in the wrapped phase recovered by phase-shifting demodulation and then in the absolute phase after phase unwrapping. A subsequent phase-to-depth conversion is used to recover 3D shape.

In conventional FPP, depth is often recovered through geometric calibration of the camera--projector pair, typically by modeling both devices as pinhole systems with lens distortion\cite{BU2023112996}. Camera calibration is well studied\cite{SALVI20021617, faugeras1993three, 888718}, whereas projector calibration is more difficult because the projector cannot observe the scene directly. An alternative family of methods avoids explicit projector calibration by learning a direct mapping from phase to depth at each camera pixel, often using lookup tables (LUTs), low-order polynomials, or rational functions \cite{Vargas:23, Son:25}. In the structured-light literature, these methods are often described as pixel-wise or model-free calibration. In this paper, we use that term in the narrower sense of avoiding an explicit calibrated projector model while still treating the camera as a calibrated imaging device.

A practical limitation of pixel-wise calibration is that each pixel is fitted independently. Neighboring pixels can therefore acquire substantially different phase-to-depth functions even when they observe the same smooth surface. This behavior is especially visible for higher-order polynomial or rational fits, whose coefficient maps can fluctuate strongly from pixel to pixel and introduce structured artifacts in reconstructed surfaces.

A second practical issue is the coordinate system used for calibration. Absolute phase is naturally computed on the undistorted sampling grid of the reference camera; we refer to this grid as the native grid. By contrast, external depth labels acquired from external sources, such as active stereo, are typically produced in rectified coordinates. Calibration therefore requires phase and depth to be expressed on a common sampling grid. A common solution is to rectify the phase maps so that they match the rectified depth labels. In this work, we instead map the depth supervision onto the native grid and construct calibration pairs directly in the reference-camera coordinates. We refer to this choice as native-grid pairing. This avoids rectifying the phase maps and preserves the camera-grid representation used by the proposed formulation.

To address these two issues, we propose a phase--depth calibration whose spatial dependence is shared across the image rather than fitted independently at each pixel. We refer to this as a \emph{spatially coupled} model. The shared mapping is estimated from multiple planar calibration captures and, when needed, is augmented by a bounded, spatially smooth correction field that captures small repeatable residual effects not explained by the shared model. 

\begin{figure*}[tp]
\centering
\includegraphics[width=\linewidth]{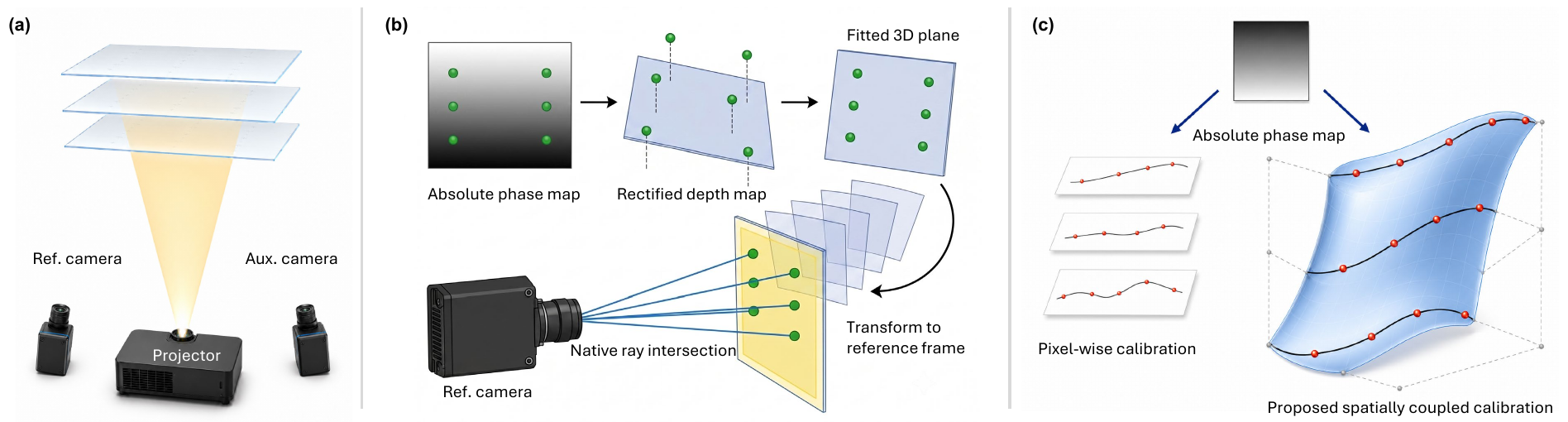}
\caption{Overview of the proposed pipeline. (a) Calibration setup: projector, reference camera, auxiliary stereo camera, and a planar target translated through multiple depths. (b) Native-grid pairing: a plane fitted to rectified stereo depth is transferred to the reference-camera frame and intersected with native camera rays to produce native-grid depth labels \(z_{n,uv}\), avoiding phase rectification. (c) Pixel-wise fitting estimates an independent \(f_{uv}(\Phi)\) per pixel, whereas the proposed method shares a base mapping \(T(u,v,\Phi)\) with a bounded smooth residual \(R(u,v)\).}
\label{fig:schematic}
\end{figure*}

The main contributions of this paper are as follows: 1) We formulate a spatially coupled phase--depth model in which all pixels share a single low-dimensional transformation---global phase scalars with affine spatial terms on the camera grid---instead of independent per-pixel functions, together with a regularized framework that jointly estimates this shared mapping and a bounded, spatially smooth correction field from planar calibration data. 2) We propose a practical native-grid pairing pipeline that constructs phase--depth calibration pairs from external supervision, such as active stereo, by fitting planes in 3D and sampling them back onto the reference-camera grid, without rectifying the phase maps. 3) We experimentally validate the proposed method on a dental target with high-resolution scanner ground truth, showing improved spatial coherence relative to conventional pixel-wise polynomial and rational calibration, with accuracy competitive with an active-stereo reference under our evaluation setup. An overview of the system setup is shown in Fig.~\ref{fig:schematic}.

\section{Principles}\label{sec:principles} 

\subsection{Absolute Phase Map Retrieval}\label{sec:phase_retrieval}
\begin{figure}[tbp]
\centering

\begin{overpic}[width=\linewidth]{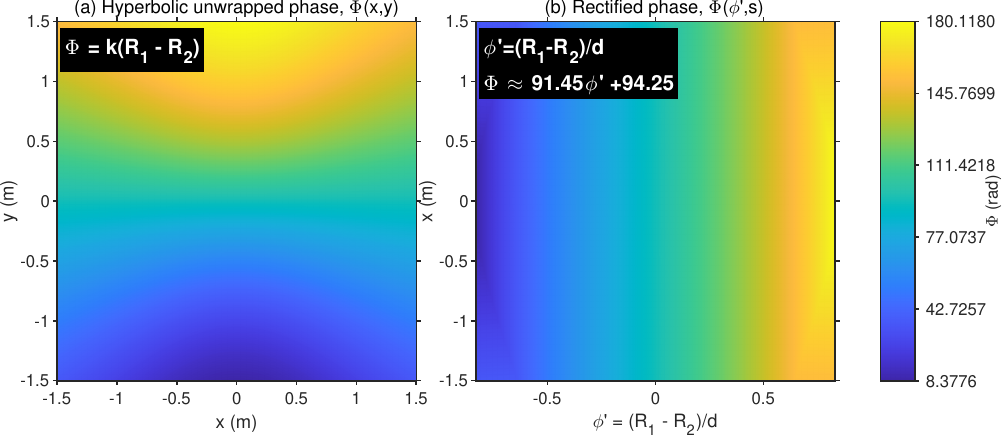}
  \put(2,43){\footnotesize\textbf{(a)}}
  \put(43,43){\footnotesize\textbf{(b)}}
\end{overpic}
\caption{Hyperbolic-phase reparameterization. (a) Raw absolute phase field \(\Phi(x,y)\) under a hyperbolic fringe parameterization. (b) After the monotonic reparameterization \(\phi'=(R_1-R_2)/d\), the phase becomes approximately affine. This illustrates the projector-model-free scope only; the experiments use standard phase-shifted fringes.}
\label{fig:phase_reparam}
\end{figure}

We work with absolute phase defined on the undistorted camera pixel grid $(u,v)$. For an ideal planar target, the absolute phase field is expected to be smooth and, under the operating geometry considered here, well-approximated by an affine (planar) function of $(u,v)$ on the native undistorted camera grid. For non-linear fringe families (e.g., circular or hyperbolic patterns), a monotonic reparameterization of the recovered phase index can be applied so that the resulting phase field over a plane becomes approximately affine. Fig.~\ref{fig:phase_reparam} illustrates such a reparameterization for a hyperbolic pattern pair.

The proposed calibration is agnostic to the specific fringe family; we describe an $N$-step phase-shifting implementation for concreteness. Under standard phase-shifting profilometry \cite{ZUO201823}, the captured intensity at pixel $(u,v)$ in the $n$-th exposure is modeled as
\begin{equation}
\label{n_step}
I_{n,uv} = I'_{uv} + I''_{uv}\cos\!\left[\phi_{uv}-\frac{2\pi n}{N}\right],
\end{equation}
where $I'_{uv}$ denotes the average intensity, $I''_{uv}$ the modulation amplitude, and $\phi_{uv}$ the wrapped phase. The wrapped phase is obtained via
\begin{equation}
\label{phase_extract}
\phi_{uv} = \tan^{-1}\!\left(\frac{\sum_{n=0}^{N-1} I_{n,uv}\sin(2\pi n/N)}{\sum_{n=0}^{N-1} I_{n,uv}\cos(2\pi n/N)}\right).
\end{equation}

To obtain a continuous phase field, we unwrap $\phi_{uv}$ by estimating an integer fringe order $k_{uv}$ and forming the absolute phase
\begin{equation}
\label{fringe_order}
\Phi_{uv} = \phi_{uv} + 2\pi k_{uv}.
\end{equation}

In the experiments, $k_{uv}$ is obtained using temporal Gray-code-assisted unwrapping \cite{ZHANG2010149}, but any absolute phase retrieval method (temporal, spatial, or hybrid) can be used\cite{An:16}.

When a phase reparameterization is required by the projected pattern family, we apply a monotonic mapping $\tilde{\Phi}_{uv}=g(\Phi_{uv})$ prior to calibration, where $g(\cdot)$ is chosen such that $\tilde{\Phi}_{uv}$ is approximately affine on planar targets (Fig.~\ref{fig:phase_reparam}). In the following, we use $\Phi$ to denote the absolute phase used for calibration (either $\Phi$ or $\tilde{\Phi}$, depending on the pattern family).

\subsection{Pixel-wise Phase--Depth Calibration}\label{sec:pixelwise}
Pixel-wise calibration estimates an independent phase--depth transfer function at each camera pixel. Given a set of calibration captures indexed by $n$, one obtains paired samples $\{(\Phi_{n,uv}, z_{n,uv})\}$ at each pixel $(u,v)$, where $z_{n,uv}$ denotes depth expressed in the chosen camera coordinate convention (Section~\ref{sec:coords}). Pixel-wise calibration models
\begin{equation}
z_{uv} = f_{uv}(\Phi_{uv}),
\end{equation}
with $f_{uv}(\cdot)$ fitted independently for each $(u,v)$ using the available samples. In the structured-light literature, $f_{uv}$ is commonly implemented as a lookup table (LUT), a low-order polynomial, or a 1D rational function. 

As a representative example, a third-order polynomial model takes the form
\begin{equation}
\label{third_order}
f_{uv}(\Phi_{uv}) = c_{0,uv} + c_{1,uv}\Phi_{uv} + c_{2,uv}\Phi_{uv}^{2} + c_{3,uv}\Phi_{uv}^{3},
\end{equation}
where the coefficient fields $\{c_{i,uv}\}$ are defined over the camera pixel grid. Higher polynomial orders or rational models can further reduce training misfit on the calibration samples, but the resulting coefficient fields may exhibit strong pixel-to-pixel variation because each pixel is fitted independently. This spatial incoherence can produce spatially inconsistent depth maps and structured surface artifacts even when per-pixel training errors are small.

At runtime, reconstruction proceeds by computing an absolute phase map $\Phi$ and evaluating $f_{uv}(\Phi_{uv})$ at each pixel. Compared with global geometric calibration models, pixel-wise methods typically require storing multiple parameters per pixel (or a per-pixel LUT), increasing memory footprint and making the learned mapping sensitive to noise and local fitting variability across the image plane.

\subsection{Coordinate Frames and Notation}\label{sec:coords}
Pixel-wise calibration in Section~\ref{sec:pixelwise} requires phase and depth samples defined on a common sampling domain. In practice, absolute phase \(\Phi\) is retrieved on the undistorted reference-camera grid, whereas depth supervision \(z\) is often produced by an external sensor (e.g., active stereo) in rectified coordinates. This section defines the coordinate conventions used throughout the paper and explains why a pairing step is required to place phase and depth on a common grid for calibration.

Let $(u_d,v_d)$ denote the integer pixel coordinates in the raw (distorted) camera image. We undistort the camera using a standard calibrated distortion model so that measurements are expressed on an undistorted coordinate grid $(u,v)$, where $(u,v)$ may be non-integer and represent sampling locations in the undistorted image plane. We treat the undistorted camera as a calibrated pinhole device with intrinsics $K$; the corresponding 3D camera coordinate frame is denoted by $C$, with 3D points $\mathbf{X}_C=[x_C,y_C,z_C]^\top$ and depth defined as the axial coordinate $z \equiv z_C$. The planarity assumption used later is stated with respect to this undistorted camera coordinate system: for planar targets, the recovered absolute phase field $\Phi_{uv}$ is expected to be smooth and well-approximated by a planar (affine) function of $(u,v)$ under typical FPP geometries (Section~\ref{sec:planarity_model}).

In the experiments, dense depth supervision is obtained from an active-stereo pipeline. Such pipelines typically compute disparity and depth on a rectified image pair, where epipolar lines are horizontal and correspondences are searched along scanlines \cite{hirschmueller2008stereo}. We denote rectified image coordinates by $(u_r,v_r)$ and the associated rectified 3D frame by $R$. For rectification with respect to a chosen reference camera (here, the left camera), the 3D relationship between frames is a rotation, $\mathbf{X}_R = R_1\,\mathbf{X}_C$, where $R_1$ is the rectification rotation. However, the mapping between the native undistorted image grid $(u,v)$ and the rectified grid $(u_r,v_r)$ is a 2D resampling operation that involves interpolation and is generally not bijective in the discrete image domain \cite{786928}. Consequently, depth maps produced in the rectified domain are not naturally defined at the same sampling locations as the phase map $\Phi_{uv}$.

A common approach for pixel-wise calibration with stereo supervision is to rectify the phase maps so that phase and depth are paired in rectified coordinates \cite{SON2024111369}. While this yields valid pairs $(\Phi_{n,u_r v_r}, z_{n,u_r v_r})$, it is not the most natural representation for the present formulation. In particular, the planar structure of phase over a physical plane is most naturally expressed in the native undistorted camera coordinates, whereas rectification is an image-domain warp that can weaken that structure and requires an additional resampling step during reconstruction.

Throughout the paper, depth is reported as \(z \equiv z_C\) in the reference-camera frame \(C\). Because the proposed formulation is defined on the undistorted reference-camera grid, we ultimately construct calibration pairs \((\Phi_{n,uv}, z_{n,uv})\) on that grid. Their construction from rectified stereo supervision is described in Section~\ref{sec:native_pairing}.

\section{Proposed spatially coupled phase--depth model}\label{sec:method}
\subsection{Shared Phase--Depth Model for Planar Calibration Targets}\label{sec:planarity_model}
Our objective is to estimate a structured phase--depth mapping on the undistorted camera grid \((u,v)\). The model combines a shared image-wide mapping, estimated from planar calibration data, with an optional bounded, spatially smooth correction field. The planar calibration targets motivate the form of the shared mapping, while the correction field captures small repeatable deviations that are not explained by that shared model. This is what we mean by a \emph{spatially coupled} formulation: pixels are linked through common image-wide parameters rather than fitted independently.

For a calibrated pinhole camera, the image of any physical plane induces a 2D projective mapping (homography) in undistorted image coordinates \cite{hartley2004multiple}. If the projector is modeled as an inverse pinhole device, then the mapping from camera coordinates to the projector image is also a homography \cite{Li:15}. Consider a world plane $P$ used as a calibration target. Let $s$ denote the projector's 1D coordinate along the fringe direction (equivalently, the continuous fringe index). Restricted to $P$, $s$ is a projective function of the camera coordinates:
\begin{equation}
\label{homography_s}
s_{uv}=\frac{\mathbf{a}^\top [u, v, 1]^\top}{\mathbf{b}^\top [u, v, 1]^\top}.
\end{equation}
Since the recovered absolute phase $\Phi$ is a monotonic reparameterization of the fringe index (Section~\ref{sec:phase_retrieval}), $\Phi_{uv}$ inherits the same projective form on planar targets up to a monotonic scaling. 

In many laboratory FPP configurations, the phase field over planar targets is observed to be well approximated by an affine (planar) function of $(u,v)$ in undistorted camera coordinates, i.e., $\Phi_{uv}\approx \alpha u + \beta v + \gamma$.
This behavior is exact for affine imaging configurations (e.g., telecentric or near-fronto-parallel/narrow-FOV regimes where the denominator in Eq.~\eqref{homography_s} is approximately constant) and is a useful approximation when the working distance is large compared to the effective focal length and the measurement volume is relatively shallow.

When the projector departs from a pinhole model, the plane-induced mapping need not be exactly projective in $(u,v)$. We therefore use these projective arguments to motivate the form of the base mapping, not as a requirement that the projector obey an exact pinhole model. The formulation only requires that the calibration phase---directly or after a monotonic reparameterization (Section~\ref{sec:phase_retrieval})---be approximately planar over planar targets within the calibration volume and pose range used here.

The above constraint motivates a planarity-constrained base mapping whose structure is induced by planar calibration targets. Under a pinhole (or near-pinhole) camera--projector geometry, the depth $z$ of a 3D point in the reference camera frame can be expressed as a perspective (linear-fractional) function of image location and phase:
\begin{equation}
\label{perspective_global}
z(u,v)=T(u,v,\Phi)=\frac{\mathbf{p}^\top [u, v, \Phi, 1]^\top}{\mathbf{q}^\top [u, v, \Phi, 1]^\top},
\end{equation}
where $\mathbf{p}$ and $\mathbf{q}$ are constant vectors. Equation~\eqref{perspective_global} provides a compact global description of phase-to-depth conversion while allowing perspective effects. For a fixed pixel $(u,v)$, this reduces to a 1D rational dependence on phase,
\begin{equation}
\label{pixelwise_rational_structured}
z=f(\Phi)=\frac{A(u,v)+B\Phi}{C(u,v)+D\Phi},
\end{equation}
where $B$ and $D$ are global scalars shared across all pixels, while
$A(u,v)$ and $C(u,v)$ are affine functions of $(u,v)$ obtained by expanding
Eq.~\eqref{perspective_global}. Hence, although
Eq.~\eqref{pixelwise_rational_structured} is a one-dimensional rational
function of phase at any single pixel, its coefficients are not free per
pixel: they are tied across the image through a few global phase terms
($B$, $D$) and low-order affine spatial terms. This shared
parameterization---replacing independent per-pixel fitting---is the core of
the proposed coupling.

This shared structure is well behaved on planar calibration targets: if \(\Phi_{uv}\) is affine on a planar target and \(A(u,v)\) and \(C(u,v)\) are affine in \((u,v)\), then the resulting depth field \(z_{uv}\) computed by Eq.~\eqref{pixelwise_rational_structured} is a smooth projective field over the image grid. We refer to this property as \emph{plane-consistent} on planar calibration targets. Using the pairing construction of Section~\ref{sec:native_pairing}, the model can be estimated from multi-plane calibration data on a common grid. In the following, \emph{planarity-constrained} refers to the shared low-dimensional mapping induced by planar calibration targets. When the residual correction field \(R(u,v)\) is enabled, the final mapping departs from this plane-consistent model only through a bounded, spatially smooth correction.

\subsection{Native-Grid Pairing of Phase and Depth}\label{sec:native_pairing}
As discussed in Section~\ref{sec:coords}, calibration requires phase and depth to be defined on a common sampling grid. In our formulation, phase is computed on the native undistorted reference-camera grid, while stereo supervision is initially obtained in rectified coordinates. We therefore map the planar depth supervision onto the native reference-camera grid by evaluating fitted planes along native camera rays, yielding \(z_{n,uv}\) directly at the same sampling locations as \(\Phi_{n,uv}\).

In the experiments, undistortion and stereo rectification are implemented using standard computer-vision toolchains (e.g., OpenCV/MATLAB), which effectively apply an undistortion-and-warp operation to the raw images. For the derivation below, we assume the reference camera has been undistorted and can be treated as an ideal pinhole camera with intrinsics $K$ (Section~\ref{sec:hardware}). Let $C$ denote the reference camera 3D frame and $R$ the rectified 3D frame (Section~\ref{sec:coords}). Stereo depth is initially reconstructed in the rectified domain, providing a 3D point cloud $\{\mathbf{X}_R\}$ or a depth field $z_{u_r v_r}$ associated with rectified pixel coordinates $(u_r,v_r)$.

For each calibration capture indexed by $n$, we select pixels belonging to the planar target and reconstruct their 3D points in the rectified frame, yielding a set $\{\mathbf{X}_{R,i}\}_{i=1}^{M_n}$. We fit a plane
\begin{equation}
\label{plane_rectified}
P_{R,n}:\ \mathbf{n}_{R,n}^{\top}\mathbf{X}_R + d_{R,n} = 0
\end{equation}
using least squares with outlier rejection using RANSAC.

Stereo rectification with respect to the reference camera is a purely rotational transform in 3D \cite{hirschmueller2008stereo}. Hence, the fitted plane parameters can be transferred to the reference-camera frame $C$ by
\begin{equation}
\label{plane_to_camera}
\mathbf{n}_{C,n} = R_1^{\top}\mathbf{n}_{R,n}, \qquad d_{C,n}=d_{R,n},
\end{equation}
where $R_1$ is the rectification rotation (Section~\ref{sec:coords}). The sign of $d_{C,n}$ follows the chosen plane-normal convention; in practice, we enforce $\mathbf{n}_{C,n}$ to face the camera so that ray--plane intersection yields positive depth.

Let $\mathbf{u}=(u,v)$ denote a sampling location on the undistorted reference image. Under the pinhole model, the back-projected ray direction associated with $\mathbf{u}$ is
\begin{equation}
\label{camera_ray}
\mathbf{r}(\mathbf{u}) \ \propto\ K^{-1}[u, v, 1]^{\top},
\end{equation}
expressed in the reference-camera frame $C$ (and normalized to unit length if desired). The intersection point between this ray and the plane $P_{C,n}$ is
\begin{equation}
\label{ray_plane_intersection}
\mathbf{X}^{\star}_{C,n}(\mathbf{u})
= \lambda^{\star}_{n}(\mathbf{u})\,\mathbf{r}(\mathbf{u}), \qquad
\lambda^{\star}_{n}(\mathbf{u}) = -\frac{d_{C,n}}{\mathbf{n}_{C,n}^{\top}\mathbf{r}(\mathbf{u})}.
\end{equation}
The depth value for pixel $(u,v)$ at pose $n$ is then defined as
\begin{equation}
\label{native_depth}
z_{n,uv} \equiv \left[\mathbf{X}^{\star}_{C,n}(\mathbf{u})\right]_z,
\end{equation}
where $[\cdot]_z$ denotes the $z$-component in the reference-camera frame. This produces the native-grid depth field $z_{n,uv}$ required for calibration.

Finally, the phase map $\Phi_{n,uv}$ is computed from the phase-shifted images on the same undistorted grid (Section~\ref{sec:phase_retrieval}). The paired samples $(\Phi_{n,uv}, z_{n,uv})$ are thus obtained in a consistent coordinate system and are used to estimate the structured phase--depth model in Section~\ref{sec:model_forms}--Section~\ref{sec:impl_profile}. The native-grid pairing procedure is illustrated schematically in Fig.~\ref{fig:schematic}, which summarizes the conversion from rectified stereo plane supervision to native-grid depth labels on the reference camera.

The ray--plane intersection in Eq.~\eqref{ray_plane_intersection} is valid when the ray intersects the observed planar region and when $\mathbf{n}_{C,n}^{\top}\mathbf{r}(\mathbf{u})$ is sufficiently far from zero (i.e., the plane is not grazing with respect to the viewing direction). In practice, we apply a binary mask $M_{n,uv}$ to restrict calibration to pixels for which (i) the phase is valid (sufficient modulation and successful unwrapping), and (ii) the planar target is visible and used in plane fitting in the rectified 3D reconstruction. All reported calibration losses and metrics are evaluated over the masked set $\{(n,u,v): M_{n,uv}=1\}$.

\subsection{Model Forms and Parameter Estimation}\label{sec:model_forms}
Given the paired calibration samples \((\Phi_{n,uv}, z_{n,uv})\) constructed in Section~\ref{sec:native_pairing}, we estimate a structured phase--depth model with two variants: an affine model suitable for near-linear phase--depth regimes, and a perspective (linear-fractional) model for wider ranges where projective effects are non-negligible. In both cases, the mapping is augmented by a pose-invariant correction field \(R(u,v)\) defined on the image grid. The model parameters are estimated jointly from all valid samples using a regularized objective that combines data fidelity with amplitude and smoothness constraints on \(R(u,v)\).

\subsubsection{Affine model}
In many practical measurement volumes (e.g., shallow depth range relative to the working distance), the phase--depth relationship is well approximated locally by a linear dependence on phase, with spatially varying offset due to the imaging geometry. We model depth as
\begin{equation}
\label{affine_model}
\hat z_{n,uv} = A(u,v) + B\,\Phi_{n,uv} + R(u,v),
\end{equation}
where $B$ is a global phase-to-depth slope shared across the image, $A(u,v)$ is a spatially varying offset, and $R(u,v)$ is a spatial correction field. Consistent with the planar constraint in Section~\ref{sec:planarity_model}, we parameterize $A(u,v)$ as an affine function of the image coordinates,
\begin{equation}
\label{A_affine}
A(u,v) = a_0 + a_1 u + a_2 v.
\end{equation}
The affine model thus uses a small set of global parameters $(a_0,a_1,a_2,B)$ together with a bounded correction field $R$.

\subsubsection{Perspective (linear-fractional) model}
When the depth range spans a wider interval or when perspective effects are visible across the field of view, the depth dependence on phase is better captured by a linear-fractional form. We model
\begin{equation}
\label{perspective_model}
\hat z_{n,uv} =
\frac{A(u,v) + B\,\Phi_{n,uv}}
     {C(u,v) + D\,\Phi_{n,uv}}
+ R(u,v),
\end{equation}
where $B$ and $D$ are global scalars and the spatial terms are again modeled as affine functions of $(u,v)$,
\begin{equation}
\label{AC_affine}
A(u,v)=a_0+a_1u+a_2v,\qquad C(u,v)=c_0+c_1u+c_2v.
\end{equation}
Equation~\eqref{perspective_model} reduces to a 1D rational dependence on phase at each pixel, but with parameters coupled across the image through the low-dimensional spatial model. This is consistent with the fact that in stereo geometry, depth is a rational function of disparity, and disparity is monotonic in phase under conventional fringe encoding.

In the perspective model, the denominator can in principle approach zero outside the valid measurement volume. To stabilize estimation, we initialize the denominator near the identity form (Section~\ref{sec:impl_profile}) and optionally apply a mild denominator reparameterization during fitting, as described in Section~\ref{sec:correction_field}.

\subsection{Correction Field, Regularization, and Constraints}\label{sec:correction_field}
The correction field \(R(u,v)\) is introduced to capture residual systematic effects that are not explained by the shared affine or perspective model, such as small deviations from ideal projective behavior, residual lens distortion after undistortion, and repeatable phase-to-depth bias patterns due to non-ideal projector behavior. Since these effects are properties of the optical configuration and coordinate convention, we model \(R(u,v)\) as constant across calibration captures \(n\) (pose-invariant), while allowing it to vary smoothly over the image grid.

To prevent the solution from relying excessively on \(R(u,v)\), and to discourage the correction field from absorbing trends already represented by the shared model, we constrain \(R(u,v)\) to be both low-amplitude and spatially smooth. We implement this by combining (i) a data fidelity term, (ii) a magnitude penalty on \(R(u,v)\), and (iii) a smoothness penalty on \(R(u,v)\). For the affine model, the objective takes the form
\begin{equation}
\label{objective_affine}
\min_{a_0,a_1,a_2,B,R}\;
\sum_{(n,u,v)\in\Omega} \rho\!\left(z_{n,uv}-\hat z_{n,uv}\right)
+\lambda_R\,\|R\|_2^2
+\lambda_S\,\|\nabla R\|_2^2,
\end{equation}

with $\hat z_{n,uv}$ given by Eq.~\eqref{affine_model}. The same form applies to the perspective model by substituting Eq.~\eqref{perspective_model}. Here, $\rho(\cdot)$ is typically the squared loss (L2), although robust losses such as Huber/Tukey loss can be used if outliers are present. We use squared loss because gross outliers are removed prior to optimization via the validity mask $M_{n,uv}$ (phase validity, stereo confidence, and plane inliers). With $R$ explicitly bounded and regularized, the remaining residuals are dominated by small, approximately zero-mean measurement noise. Robust losses were therefore not required in our setting.

In addition to these penalties, we bound the correction field to maintain a physically interpretable correction magnitude,
\begin{equation}
\label{R_cap}
|R(u,v)| \le R_{\max}\quad \forall(u,v),
\end{equation}
implemented either as a hard projection/clipping step or as a soft barrier penalty.
The weights $\lambda_R$ and $\lambda_S$ and the bound $R_{\max}$ are selected to trade off local fidelity against spatial coherence, and are reported with the implementation details (Section~\ref{sec:impl_profile}). When $R$ is fixed to 0, the mapping reduces to a fully specified low-dimensional model with no regularization hyperparameters. The $R \neq 0$ variants introduce controlled flexibility to capture repeatable bias while preserving spatial coherence.

For the perspective model, we additionally discourage denominator collapse during fitting by reparameterizing the denominator as
\begin{equation}
\label{denom_stable}
C(u,v) + D\,\Phi_{n,uv} \;=\; 1 + \rho_d\,\big(c_0+c_1u+c_2v + D\,\Phi_{n,uv}\big),
\end{equation}
with a small $\rho_d$ (e.g., $10^{-2}$), which keeps the denominator close to unity during early iterations and improves numerical stability. Gradient clipping and early stopping criteria are applied as needed; implementation details are given in Section~\ref{sec:impl_profile}.

\subsection{Implementation and Computational Profile}\label{sec:impl_profile}
This section summarizes practical aspects of parameter estimation and runtime evaluation. Since calibration is performed offline, our primary computational consideration is the runtime cost and memory footprint during reconstruction, rather than the wall-clock time required for the optimizer to converge.

Model parameters are estimated by minimizing the regularized objectives in Eqs.~\eqref{objective_affine}--\eqref{denom_stable} over the valid sample set defined by $M_{n,uv}$ (Section~\ref{sec:native_pairing}). In all experiments, optimization is performed using a quasi-Newton method (L-BFGS) implemented in PyTorch. We found L-BFGS to converge reliably for both affine and perspective variants when initialized near the affine regime, although other first-order optimizers (e.g., Adam) can also be used.

For the perspective model, the denominator parameters are initialized to zero. Under the reparameterization of Eq.~\eqref{denom_stable}, this yields a denominator of unity (an identity mapping), which avoids early-stage denominator collapse. $\rho_d$ is set to a small constant. Gradient clipping and a maximum-iteration stopping rule are applied to ensure numerical stability. Unless otherwise stated, optimizer settings are kept at their library defaults when not exposed for configuration.

$\lambda_R$ and $\lambda_S$ are selected to enforce a low-amplitude, slowly varying correction field in $R(u, v)$; rather than tuning for minimum training misfit, we report the resulting magnitude and smoothness for $R$ and include results with $R = 0$ to isolate the contribution of the correction term. For the rest of the experiments, $\rho_d = 10^{-2}$, $R_{\max} = 2.0 \mathrm{\mu m}$, $\lambda_R = 0.001$, $\lambda_S = 1$ was used throughout.

The affine model admits closed-form least-squares initialization. In principle, the global slope $B$ and the spatial offset parameters $(a_0,a_1,a_2)$ can be obtained from a two-stage linear regression using the calibration samples $(\Phi_{n,uv}, z_{n,uv})$, and the correction field initialized as $R(u,v)=0$.

During reconstruction, the calibrated mapping is evaluated point-wise on the image grid using the recovered phase map $\Phi$. In the affine case (Eq.~\eqref{affine_model}), we precompute the spatial offset $A(u,v)$ and optionally absorb the correction field into a single offset map,
\begin{equation}
\label{precompute_affine}
A'(u,v) = A(u,v) + R(u,v),
\end{equation}
so that runtime evaluation reduces to
\begin{equation}
\label{runtime_affine}
z(u,v) = A'(u,v) + B\,\Phi(u,v),
\end{equation}
which is a small number of element-wise operations.

In the perspective case (Eq.~\eqref{perspective_model}), we precompute $A(u,v)$ and $C(u,v)$ as affine maps over the pixel grid and evaluate
\begin{equation}
\label{runtime_persp}
z(u,v)=\frac{A(u,v)+B\,\Phi(u,v)}{C(u,v)+D\,\Phi(u,v)} + R(u,v),
\end{equation}
using element-wise tensor operations. The correction field $R(u,v)$ cannot generally be absorbed into a single precomputed offset in the perspective model, but it still contributes only one additional element-wise addition.

We report the arithmetic cost of the proposed mappings in terms of per-pixel floating-point operations (FLOPs), since end-to-end throughput in modern 3D reconstruction pipelines is often limited by memory bandwidth and data movement rather than raw arithmetic. For reference, a conventional camera--projector triangulation pipeline typically requires solving a small linear system per pixel (or an equivalent sequence of matrix operations), whereas the proposed model evaluates a fixed set of tensor operations per pixel. The resulting FLOP counts are summarized in Table~\ref{tab:impl}. 

We also report the memory footprint of each calibration model. Pixel-wise polynomial and rational calibration store multiple parameters per pixel (or per-pixel LUT entries), resulting in a parameter footprint that scales with image resolution and model order. In the proposed model, the global affine/perspective parameters contribute negligible memory, while the dominant resolution-dependent storage is the correction field $R(u,v)$ when enabled. Table~\ref{tab:impl} summarizes parameter counts and approximate storage requirements for all compared methods.

\begin{table*}[tp]
\centering
\caption{Implementation details and computational profile for reconstruction.}
\label{tab:impl}
\setlength{\tabcolsep}{6pt}
\begin{tabular}{@{}p{4.1cm} p{1.4cm} p{1.0cm} p{2.3cm} p{2.5cm} p{1.0cm} p{3.5cm}@{}}
\toprule
\textbf{Method} &
\textbf{Per-pixel params} &
\textbf{Global params} &
\textbf{Precomputed maps} &
\textbf{Approx. Ops per-pixel} &
\textbf{Rect. frame} &
\textbf{Notes}\\
\midrule
Active stereo  &
0  &
$P_c$ &
$P_c$ &
Stereo matching &
Y &
Bandwidth-bound\\

Pinhole triangulation &
0  &
$P_c,P_p$ &
$P_c, P_p$ &
Triangulation & 
Y &
Requires projector cal. \\

Pixel-wise poly (order $k$) &
$k{+}1$ floats &
None &
None &
2k & 
Y &
Potential discontinuities\\

Pixel-wise 1D rational  &
$4$ floats &
None &
None &
$\approx$3 FLOPs &
Y &
Potential singularities\\

Proposed affine ($R=0$) &
0 &
4 floats &
$A(u,v)$ &
$\approx$2 FLOPs &
N &
Element-wise\\

Proposed affine ($R\neq0$) &
1 float ($R$) &
4 floats &
$A'(u,v)=A+R$ &
$\approx$2 FLOPs &
N &
Dominant storage: $R$\\

Proposed perspective ($R=0$) &
0 &
8 floats &
$A(u,v),C(u,v)$ &
$\approx$5 FLOPs &
N &
Element-wise\\

Proposed perspective ($R\neq0$) &
1 float ($R$) &
8 floats &
$A(u,v),C(u,v)$ &
$\approx$6 FLOPs &
N &
+$R$ add\\
\bottomrule
\end{tabular}
\end{table*}

\section{Experimental protocol and reporting conventions}\label{sec:protocol}

\subsection{Hardware setup and geometric calibration}\label{sec:hardware}
All experiments were conducted using an active-stereo structured light system consisting of a calibrated stereo camera pair and an uncalibrated projector. One camera is designated as the reference camera used to index the absolute phase map and to define the native camera coordinate frame $C$ (Section~\ref{sec:coords}). The projector is operated as a pattern emitter only; no pinhole projector model is assumed by the proposed calibration.

The system uses two monochrome cameras (uEye UI304xCP-M, 480$\times$ 528) mounted with synchronized acquisition. The projector (DLP2010LC-2205, 480$\times$ 854) projects phase-shifting patterns under mild defocus to suppress high-frequency binary artifacts. Unless otherwise stated, all phase processing and calibration are performed on the undistorted reference-camera sampling grid at a resolution of \textit{$480\times 528$}.

The reference camera is calibrated using a standard pinhole model with lens distortion \cite{zhang2009projector}. We use an asymmetric \SI{1}{mm} checkerboard target with $54$ detected grid points and capture $25$ calibration poses. Intrinsic parameters are estimated via Zhang's method, and each pose's extrinsics are refined using a PnP-based procedure. The mean reprojection error of the single-camera calibration is $0.032 \mathrm{px}$. Using the estimated intrinsics and distortion coefficients, each distorted pixel $(u_d,v_d)$ is mapped to an undistorted sampling location $(u,v)$, and all phase maps are subsequently computed and stored on this undistorted grid (Section~\ref{sec:phase_retrieval}). We did not propagate calibration uncertainty through the full reconstruction chain; reprojection errors are reported as a proxy for geometric calibration quality.

For the active-stereo baseline and for generating depth supervision during calibration, the stereo camera pair is calibrated using standard stereo calibration to estimate the relative rotation and translation between cameras, followed by stereo rectification \cite{hirschmueller2008stereo}. The stereo calibration yields a mean reprojection error of $0.039 \mathrm{px}$. Rectification parameters are computed with respect to the reference camera so that rectified epipolar lines align horizontally. For clarity, our mathematical derivations treat the reference camera as undistorted and pinhole (Section~\ref{sec:coords}); calibration is performed on native-grid phase--depth pairs using the pairing construction of Section~\ref{sec:native_pairing}.

Unless otherwise stated, all reported depth values are expressed as the $z$-coordinate in the reference camera frame $C$ (Section~\ref{sec:coords}). Additional details on the generation of planar reference depth from stereo reconstructions and plane fitting are provided in Section~\ref{sec:ref_3d}.

\subsection{Reference 3D Data Generation for Calibration}\label{sec:ref_3d}
This section summarizes the generation of reference depth labels for the planar calibration captures only. For each pose, the rectified stereo reconstruction is used to estimate the corresponding 3D plane; native-grid depth labels \(z_{n,uv}\) and the validity mask \(M_{n,uv}\) are then obtained by the pairing construction of Section~\ref{sec:native_pairing}. 

Absolute phase is recovered from a six-step phase-shifted sequence (Section~\ref{sec:phase_retrieval}). For each calibration pose, we capture the full pattern sequence for phase retrieval and a corresponding stereo pair for depth reconstruction. We acquire $40$ planar calibration captures spanning a depth range of approximately $14 \mathrm{mm}$ within the usable measurement volume, with the target maintained at approximately fixed orientation relative to the reference camera while translated through depth. In our experiments, the planar target orientation was held approximately fixed across calibration captures so that variations in phase quality, stereo support, and the validity mask were driven primarily by depth rather than by pose-dependent illumination or foreshortening effects.

For each calibration pose $n$, a rectified stereo pair is reconstructed using the active-stereo pipeline (Section~\ref{sec:hardware}), producing either a rectified depth map $z_{n,u_r v_r}$ or a rectified 3D point cloud $\{\mathbf{X}_{R,i}\}$. Points with invalid disparity, low confidence, or saturation are discarded.

A geometric plane is fitted directly in the reconstructed 3D coordinates (not in image coordinates) by least squares with RANSAC, yielding the plane parameters $(\mathbf{n}_{R,n}, d_{R,n})$ for each calibration pose. These are then transferred to the reference-camera frame and converted to native-grid depth labels by the pairing construction of Section~\ref{sec:native_pairing}.

\subsection{Metrics and Reporting Conventions}\label{sec:metrics}
We report results for three categories: (i) calibration diagnostics on planar data, (ii) reconstruction performance on freeform dental targets using an external ground-truth (GT) scanner, and (iii) spatial interpolation tests. Because different methods naturally produce depth in different coordinate systems (native undistorted grid vs rectified stereo grid vs scanner coordinates), we distinguish between 2D per-pixel metrics (when available) and 3D surface metrics (used for cross-system comparisons).

All depth values are reported in millimeters, with micrometer-scale errors reported in $\si{\micro\meter}$ for readability. Unless otherwise stated, depth $z$ refers to the $z$-coordinate in the reference camera frame $C$ (Section~\ref{sec:coords}). For calibration, all losses and metrics are computed on the masked set $\{(n,u,v):M_{n,uv}=1\}$, where $M_{n,uv}$ encodes valid phase, planar visibility, and valid stereo support (Section~\ref{sec:native_pairing} and Section~\ref{sec:ref_3d}).

For phase-to-depth model fit error, highly flexible per-pixel models drive the training misfit to numerical precision; therefore, misfit alone is not used as the primary indicator of evaluating fit or spatial consistency.

When the correction field $R(u, v)$ is enabled, its effects are summarized by its peak-to-peak amplitude, RMS value, and spatial smoothness proxy. We visualize $R(u,v)$ as a 2D map later in Fig.\ref{fig:Rmap}. For methods that produce per-pixel parameters (e.g., pixel-wise polynomial/rational fits), we report the same spatial smoothness proxy. Representative coefficient maps are visualized in Fig.\ref{fig:coeffmap} to compare spatial coherence.

Freeform reconstruction is evaluated using a high-resolution 3D scanner as ground truth. Since the scanner data and the reconstructed surfaces exist in different coordinate frames, direct per-pixel depth comparison is generally not meaningful. Instead, each reconstruction is converted to a triangle mesh and compared to the scanner mesh via 3D registration and point-to-surface distances.

Specifically, for each dental capture $m$, we generate a reconstructed point cloud or depth map, convert it to a surface mesh using normal-based surface integration, and apply rigid alignment to the scanner mesh using ICP with trimmed correspondences using the GOM Inspect 2018 software. We report the registration residual using final alignment RMSE as a sanity check for the registration quality and also report point-to-surface distance statistics over the overlapping region. These values are reported for each of the five dental views and aggregated across views.

For methods that operate on the same native undistorted grid and produce dense depth maps in the same pixel coordinates (e.g., the proposed method and the native-grid version of pixel-wise calibration), we additionally report per-pixel depth difference maps and summary statistics on common valid pixels. For baselines that operate natively in rectified coordinates, we do not force per-pixel comparisons unless a common resampling is explicitly performed, as this introduces interpolation-dependent artifacts.

To evaluate the stability of the spatially coupled model under reduced calibration coverage, we perform a spatial hold-out experiment in which the model is calibrated on a subregion of the image and evaluated on the full field. We report the same dental surface-deviation statistics as above for both $R=0$ and $R\neq 0$ configurations. 

\section{Reconstruction experiments and results}\label{sec:results}
This section evaluates the proposed spatially coupled phase--depth calibration on (i) planar calibration data for diagnostic characterization and (ii) freeform dental targets with external ground truth. All protocols, coordinate conventions, and metrics follow Section~\ref{sec:protocol}. Unless otherwise stated, results are reported in $\si{\micro\meter}$ and computed over valid pixels/regions defined by the masks in Section~\ref{sec:native_pairing} and Section~\ref{sec:metrics}.

\subsection{Evaluation Setup for Freeform Reconstruction}\label{sec:exp_setup}
This subsection introduces the additional experimental components required for freeform evaluation beyond the calibration protocol in Section~\ref{sec:protocol}: (i) the dental target and scanner ground truth, and (ii) additional reconstruction baselines.

We evaluate reconstruction on a dental model made of colored gypsum with constant surface reflection properties under constant illumination. The model is captured from five different viewpoints, each focusing on a different feature of the model with variable depth and normal distribution. A high-resolution 3D scanner (ZEISS O-INSPECT, 0.9 $\mu\mathrm{m}$ resolution) provides reference geometry for each viewpoint. 

We compare the proposed method against four baselines:
\begin{itemize}
\item \textbf{Active stereo baseline:} depth reconstructed directly from the stereo pipeline in rectified coordinates (Section~\ref{sec:hardware}). For 3D surface comparison, the stereo reconstruction is converted to a mesh and aligned to scanner GT using the same procedure as other methods (Section~\ref{sec:metrics}).
\item \textbf{Geometric triangulation baseline:} a conventional camera--projector calibration followed by triangulation using a pinhole projector model (Zhang's Method). This baseline is included as a reference for classical structured-light calibration assumptions.
\item \textbf{Pixel-wise polynomial calibration:} per-pixel phase-depth regression of order-k polynomials using QR decomposition. For this research, $k=3$ was used.
\item \textbf{Pixel-wise rational calibration:} per-pixel 1D rational mapping using QR decomposition. Singularities during reconstruction are detected as outliers and ignored during registration.
\end{itemize}
For fair comparison, all phase maps used by per-pixel methods are defined on the same undistorted reference-camera grid. When a baseline is naturally defined in rectified coordinates, we report only mesh-level metrics (Section~\ref{sec:metrics}) unless an explicit common resampling is performed.

For the proposed model, we evaluate both the affine and perspective forms, each with (i) no correction field ($R\equiv 0$) and (ii) a bounded, smooth correction field ($R\neq 0$).

\subsection{Calibration Diagnostics and Spatial Consistency}\label{sec:cal_diag}
This subsection evaluates calibration outcomes on planar data to characterize spatial consistency and the role of the correction field. 
Following the convention of Section~\ref{sec:metrics}, we emphasize diagnostics of spatial stability rather than data misfit alone.

\begin{figure}[tbp]
\centering
\begin{overpic}[width=\linewidth]{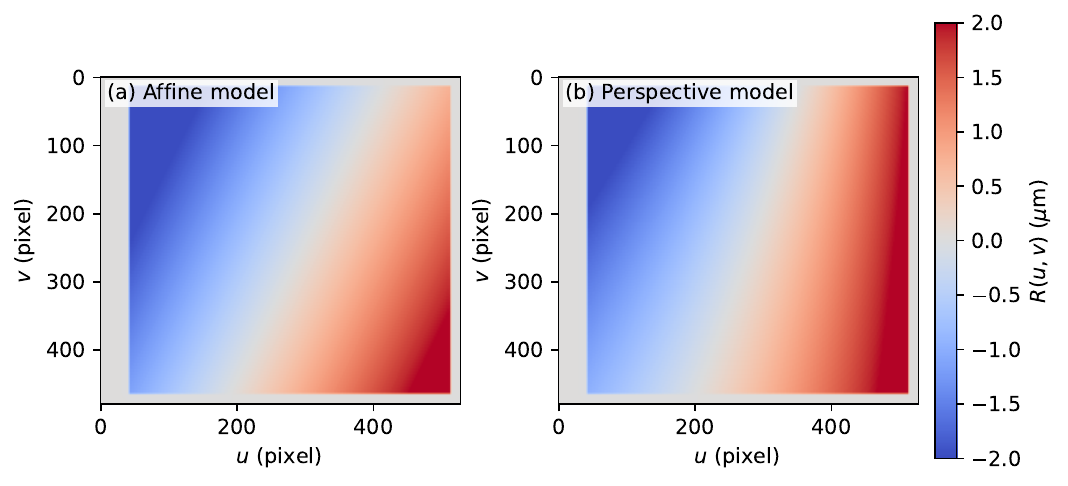}
  \put(2,40){\footnotesize\textbf{(a)}}
  \put(43,40){\footnotesize\textbf{(b)}}
\end{overpic}

\caption{Estimated $z$-component of the correction field $R(u,v)$. The shared color scale is saturated at $\pm R_{\max}=2.0\,\mu\mathrm{m}$. (a) Affine model. (b) Perspective model.}
\label{fig:Rmap}
\end{figure}

We visualize the estimated correction field $R(u,v)$ for the affine and perspective models and report summary statistics (RMS, peak-to-peak, and smoothness proxy). Representative maps are shown in Fig.~\ref{fig:Rmap}.
The magnitude of $R$ is bounded by $R_{\max} = 2 \mathrm{\mu m}$, and we verify that the learned $R$ remains well within this bound across all calibration captures.

\begin{figure}[tbp]
\centering
\includegraphics[width=0.8\columnwidth]{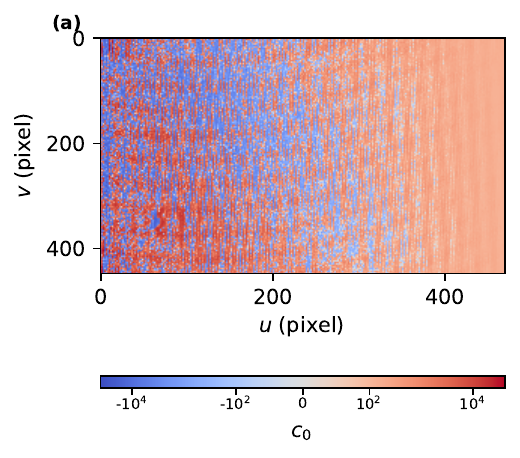}
\caption{Representative spatial parameter map of $c_{0, uv}$ illustrating pixel-to-pixel instability in per-pixel calibration of polynomial with order 3. Only contains pixels deemed valid by $M(u,v)$. }
\label{fig:coeffmap}
\end{figure}

For per-pixel models, training misfit approaches numerical precision and is therefore not interpreted as general measurement accuracy, especially for near-affine models which generally approach floating point precision. For the proposed model, training misfit decreases when $R$ is enabled, reflecting the role of the correction field in capturing repeatable bias patterns while preserving spatial structure. The model used for the experiment resulted in a training misfit of 3.344 (0.285) (R = 0) and 3.032 (0.277) (R $\neq$ 0) for the affine case, while the perspective case resulted in 1.142 (0.384) (R = 0) and 0.681 (0.251) (R $\neq$ 0). 

To relate these diagnostics to conventional per-pixel calibration, Fig.~\ref{fig:coeffmap} shows a representative coefficient map from a polynomial fit. Unlike the low-dimensional parameterization of the proposed model, per-pixel coefficient fields can fluctuate sharply across neighboring pixels, providing a direct visual explanation for the structured surface artifacts later seen on freeform reconstructions. 

\subsection{Dental Reconstruction Accuracy Against Scanner Ground Truth}\label{sec:dental_results}
We evaluate freeform reconstruction accuracy using point-to-surface deviations between reconstructed meshes and scanner GT (Section~\ref{sec:metrics}). All methods use the same phase retrieval and masking rules when applicable; only the phase-to-depth conversion differs.

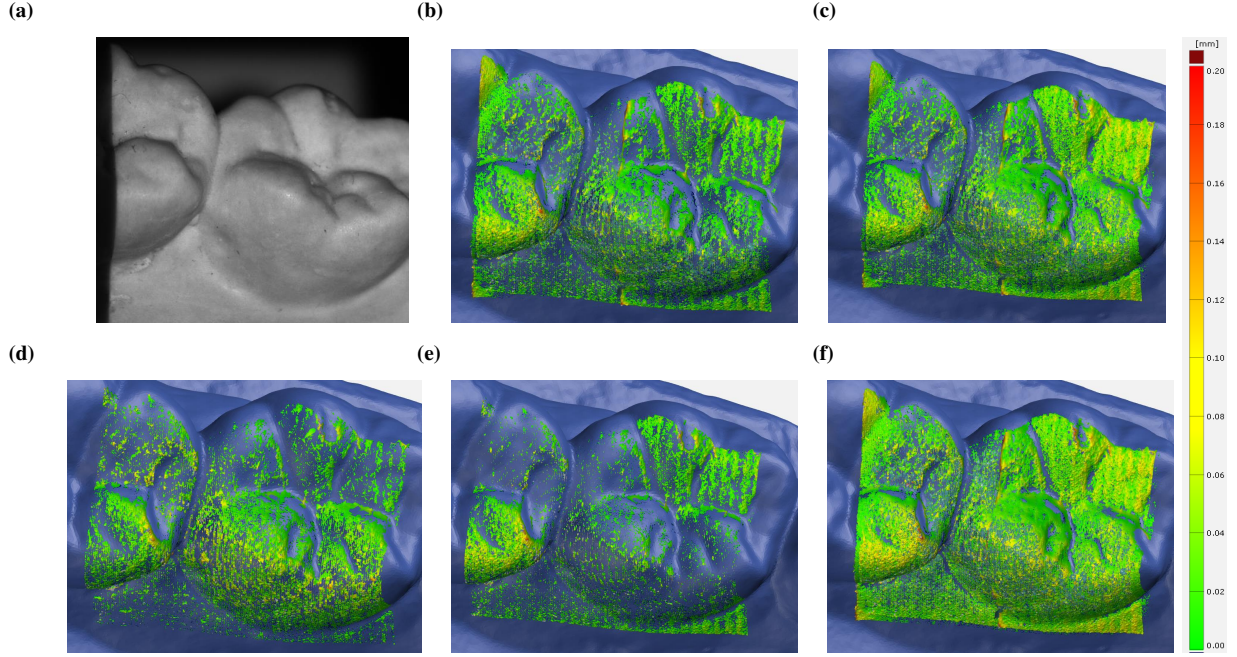
\begin{figure*}[tp]
\centering
\begin{overpic}[width=0.95\linewidth]{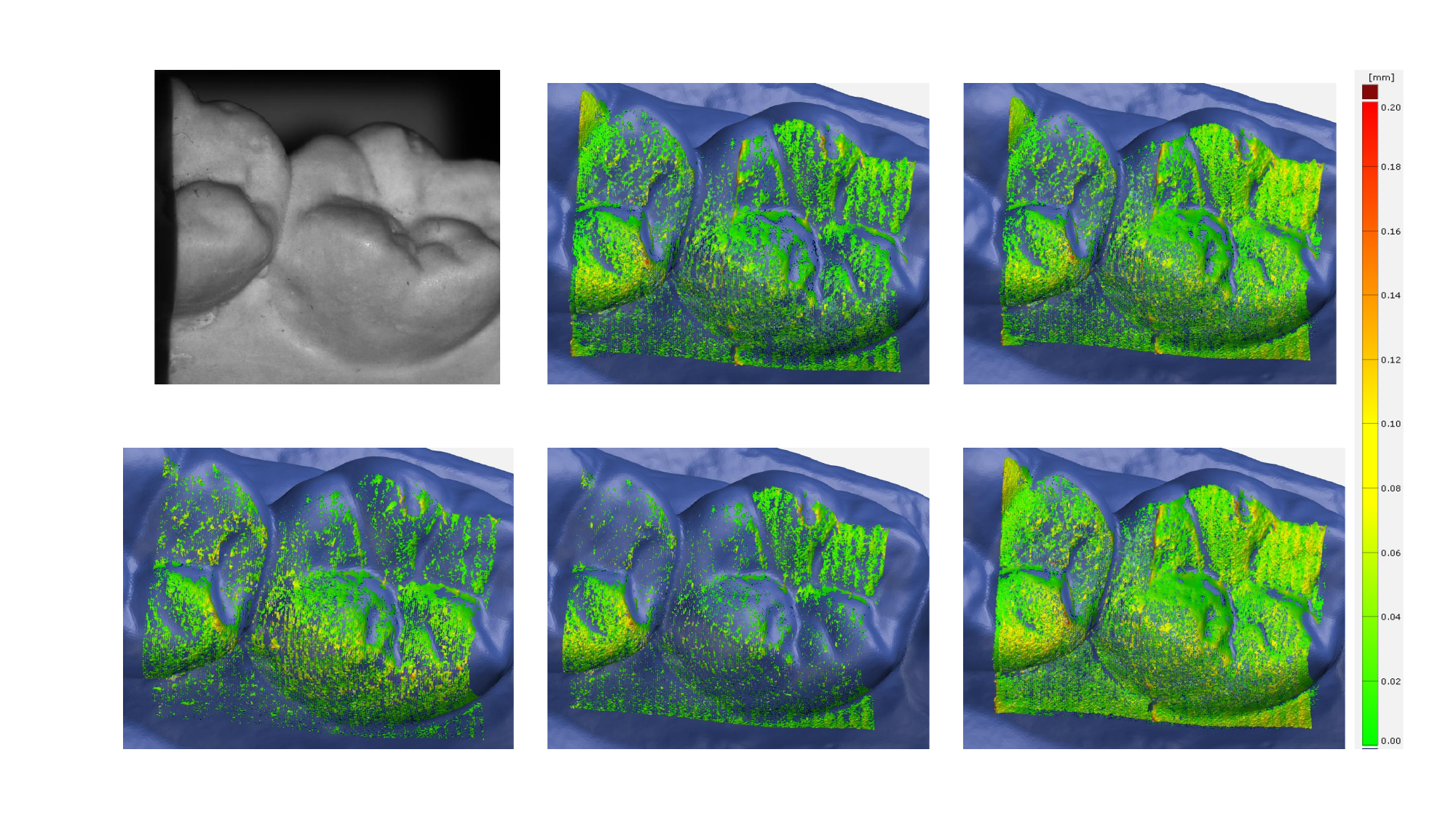}
  \put(4,53){\footnotesize\textbf{(a)}}
  \put(35,53){\footnotesize\textbf{(b)}}
  \put(65,53){\footnotesize\textbf{(c)}}
  \put(4,27){\footnotesize\textbf{(d)}}
  \put(35,27){\footnotesize\textbf{(e)}}
  \put(65,27){\footnotesize\textbf{(f)}}
\end{overpic}
\caption{Representative dental reconstructions and deviation maps against scanner ground truth, corresponding to View 4 of Table \ref{tab:dental_quant}. Maximum deviation is capped at 0.2 mm. Coverage is 92\% for rectified models and 84\% for per-pixel models. (a) Raw, (b) Proposed Affine model, (c) Proposed Perspective model, (d) Conventional Polynomial model with $k=3$, (e) Conventional Rational model with $k=1$, (f) Conventional Active Stereo. }
\label{fig:dental_vis}
\end{figure*}

Fig.~\ref{fig:dental_vis} shows 3D deviation maps for one of the views using a fixed color scale across methods along with raw image and reconstruction results. The proposed method produces smoother, spatially consistent surfaces compared to pixel-wise calibration, while maintaining comparable deviation levels to the active-stereo baseline.

Table~\ref{tab:dental_quant} summarizes deviation statistics for five dental views and an aggregate across views.

\begin{table*}[tp]
\centering
\caption{Dental reconstruction deviation against scanner ground truth. Reported values are RMSE point-to-surface distances after rigid registration. (Section~\ref{sec:metrics}). }\label{tab:dental_quant}
\begin{tabular}{lcccccc}
\toprule
Method & View 1 & View 2 & View 3 & View 4 & View 5 & Aggregate\\
\midrule
Active stereo & 11.5 & 10.9 & 10.1 & 11.4 & 9.3 & 10.64\\
Geometric triangulation & 14.1 & 12.2 & 11.9 & 12.0 & 12.5 & 12.54\\
Pixel-wise poly & 10.7 & 13.3 & 15.0 & 14.4 & 13.3 & 13.34\\
Pixel-wise rational & 19.4 & 16.4 & 14.4 & 10.2 & 11.8 & 14.44\\
Proposed affine ($R=0$) & 18.8 & 15.9 & 12.7 & 14.4 & 10.1 & 14.38\\
Proposed affine ($R\neq 0$) & 16.8 & 13.1 & 12.2 & 13.4 & 9.2 & 12.94\\
Proposed perspective ($R=0$) & 14.2 & 15.6 & 12.9 & 9.8 & 9.9 & 12.48\\
Proposed perspective ($R\neq 0$) & 13.1 & 14.0 & 12.2 & 10.4 & 9.8 & 11.90 \\
\bottomrule
\end{tabular}
\end{table*}

The proposed affine and perspective models with $R\neq 0$ improve upon pixel-wise polynomial and rational baselines and achieve accuracy comparable to active-stereo. The $R\equiv 0$ variants remain competitive in near-affine regimes, but generally underperform the $R\neq 0$ variants on freeform surfaces.

This behavior is consistent with the calibration diagnostics of Section~\ref{sec:cal_diag}: the shared spatial parameterization suppresses the coefficient-level instability visible in Fig.~\ref{fig:coeffmap}, yielding smoother reconstructed geometry while retaining competitive accuracy. In this setting, the main benefit of the proposed model is improved spatial coherence of the learned phase--depth relation rather than lower calibration-sample misfit.

\subsection{Spatial Interpolation Verification}\label{sec:interp}
This experiment evaluates whether the proposed spatially coupled model remains stable when calibration supervision is available only on a subregion of the image. Here, ``spatial interpolation'' refers to the stability of the calibrated mapping over the full native undistorted grid; we do not interpolate phase maps or depth maps themselves.

This distinction matters because conventional per-pixel calibration estimates independent parameter values only where valid labels are available. In contrast, the proposed mapping is defined directly on the native reference-camera grid through a low-dimensional dependence on \((u,v)\), optionally augmented by a bounded smooth correction field \(R(u,v)\). The experiment therefore tests whether that spatial structure yields a stable full-frame mapping under reduced calibration coverage, rather than whether missing measurements can be inferred outside the observed region.

Because active stereo provides reliable supervision only on a subset of pixels, we restrict calibration to a \(200\times 300\) subregion and exclude all other pixels from \(M_{n,uv}\). The calibrated model is then applied to full-frame dental phase maps, and the resulting depth is compared, per pixel over the common valid mask, against the reconstruction obtained under full-frame calibration. We report results for both \(R\equiv 0\) and \(R\neq 0\) variants in Table~\ref{tab:interp}. We emphasize that this metric quantifies the \emph{stability} (self-consistency) of the calibrated mapping under reduced coverage, not its accuracy against scanner ground truth. As expected, the \(R\equiv 0\) variants are the most stable (depth changes of \(0.17\)--\(0.31\,\mu\mathrm{m}\)), since they depend only on the low-dimensional shared parameters, which remain well constrained even when estimated from a subregion. The \(R\neq 0\) variants change more (\(\approx 1\,\mu\mathrm{m}\)) because the correction field is weakly constrained outside the supervised subregion; this is the cost of its additional flexibility and indicates that, under sparse calibration coverage, the regularized base mapping (\(R\equiv 0\)) is preferable.

\begin{table}[tbp]
\centering
\caption{Effect of reduced calibration coverage. Reported values summarize the per-pixel depth difference relative to full-frame calibration ($480\times 528$), computed over the common valid mask $M(u,v)$ (Section~\ref{sec:metrics}).}
\label{tab:interp}
\begin{tabular}{@{}p{2.7cm} p{1.3cm} p{1.2cm} p{2.1cm}}
\toprule
Variant & Subregion & RMSE($\Delta z$) & median(IQR) \\
\midrule
Affine ($R=0$) & $200\times 300$ & 0.17 $\mathrm{\mu m}$ & 0.17 (0.03) $\mathrm{\mu m}$\\
Affine ($R\neq0$) & $200\times 300$ & 1.03 $\mathrm{\mu m}$& 1.24 (1.44) $\mathrm{\mu m}$\\
Perspective ($R=0$) & $200\times 300$ & 0.31 $\mathrm{\mu m}$& 0.10 (0.29) $\mathrm{\mu m}$\\
Perspective ($R\neq0$) & $200\times 300$ & 1.08 $\mathrm{\mu m}$& 0.78 (1.21) $\mathrm{\mu m}$\\
\bottomrule
\end{tabular}
\end{table}
We additionally note that error typically increases near image borders for all methods. We intentionally do not exclude border regions from evaluation, since edge behavior is part of the practical operating envelope.

\section{Conclusion}\label{sec:conclusion}
We presented a spatially coupled phase--depth calibration for fringe projection profilometry in which a shared image-wide transformation is defined on the undistorted reference-camera grid and, when needed, augmented by a bounded, spatially smooth correction field. Planar calibration targets determine the form of the shared mapping, while the correction field captures small repeatable deviations from that model.

Two model variants were considered: an affine mapping appropriate for near-linear phase--depth regimes, and a perspective (linear-fractional) mapping for wider ranges where projective effects are more pronounced. To accommodate repeatable systematic effects not captured by the low-dimensional model, we introduced a bounded, spatially smooth correction field $R(u,v)$ that is invariant across calibration captures. Experimental results show that the proposed mapping yields substantially improved spatial coherence compared with conventional pixel-wise polynomial and rational calibration, while achieving reconstruction accuracy comparable to an active-stereo baseline on freeform dental targets with scanner ground truth.

A practical advantage of the method is that calibration is carried out on the undistorted reference-camera grid even when supervision is obtained from a rectified stereo pipeline. In the proposed formulation, this is achieved by mapping planar depth supervision onto the camera grid during calibration, so that phase maps do not need to be rectified. The resulting runtime mapping is then evaluated directly on that grid using only element-wise operations.

The proposed method assumes accurate reference-camera undistortion and planar calibration targets that are well described by a single plane in the stereo reconstruction used for supervision. Violations of these assumptions (e.g., residual lens distortion, imperfect planarity, or low-confidence stereo) can bias the estimated mapping. The correction field $R(u,v)$ mitigates repeatable systematic effects but is explicitly constrained in amplitude and smoothness; consequently, large unmodeled distortions or operation outside the valid measurement volume may not be fully compensated. In practice, reconstruction performance depends on the chosen parameterization and regularization strength of $R$; we therefore report results for both $R\equiv 0$ and $R\neq 0$ to make this trade-off explicit. In addition, the experimental calibration protocol varied plane depth while keeping plane orientation approximately fixed; broader pose diversity may be beneficial in other operating regimes, but was not explored here because pose-dependent changes in phase quality and valid support would confound the intended evaluation of the spatial model itself.

A second limitation concerns phase quality. During calibration, depth supervision on planes benefits from geometric fitting, whereas we do not enforce an analogous planar constraint on the measured phase field. Although fitting or denoising phase to an ideal planar field can improve training convergence and reduce calibration misfit, we found that such calibration-time phase smoothing may not transfer to freeform reconstructions and can degrade accuracy when the runtime phase noise statistics differ. As with other per-pixel mappings, the achievable reconstruction accuracy is therefore sensitive to phase SNR and to the phase-denoising strategy used at inference time. Accordingly, the role of the proposed model is not to enforce exact planarity in the recovered phase-to-depth relation, but to retain a plane-consistent base structure while absorbing repeatable departures that persist in practical reconstruction.

Overall, the proposed calibration retains the flexibility of direct phase--depth fitting while enforcing a shared spatial structure across the image, thereby improving global consistency and reducing structured artifacts in reconstructed surfaces.

\section*{CRediT authorship contribution statement}

\noindent\textbf{Sehoon Tak} Conceptualization, Methodology, Software, Validation, Investigation, Data curation, Visualization, Writing -- original draft, Writing -- review \& editing.

\noindent\textbf{Jae-Sang Hyun} Conceptualization, Methodology, Supervision, Project administration, Writing -- review \& editing.

\section*{Declaration of competing interest}
The authors declare that they have no known competing financial interests or personal relationships that could have appeared to influence the work reported in this paper.

\section*{Data availability}
Raw image data are not publicly available due to confidentiality restrictions regarding the hardware setup. Processed data supporting the findings of this study (e.g., reconstructed meshes, error statistics, and calibration parameter fields) and the scripts used to generate the reported figures are available from the corresponding author upon reasonable request.

\section*{Acknowledgements}
This research was supported by the Technology Innovation Program (Project Name: Development of AI autonomous continuous production system technology for gas turbine blade maintenance and regeneration for power generation, Project Number: RS-2025-25447257, Contribution Rate: 50\%) funded by the Ministry of Trade, Industry and Resources (MOTIR, Korea), the Culture, Sports and Tourism R\&D Program through the Korea Creative Content Agency grant funded by the Ministry of Culture, Sports and Tourism in 2024 (Project Name: Global Talent for Generative AI Copyright Infringement and Copyright Theft, Project Number: RS-2024-00398413, Contribution Rate: 30\%), and the National Research Foundation of Korea (NRF) grant funded by the Korea government (MSIT) (Project Number: RS-2026-25498577, Contribution Rate: 20\%).

\bibliographystyle{elsarticle-num}
\vfill{}
\bibliography{ref}

\end{document}